\documentclass[letterpaper, 10pt, conference]{IEEEtran}
\IEEEoverridecommandlockouts
\usepackage{cite}
\usepackage{amsmath,amssymb,amsfonts}
\usepackage{algorithmic}
\usepackage{graphicx}
\usepackage{textcomp}
\usepackage{xcolor}
\usepackage{hyperref}
\usepackage[letterpaper, left=0.75in, right=0.75in, bottom=0.75in, top=0.75in]
{geometry}

\begin{document}

\title{\vspace{0.18in}ADMFormer: An Adaptive-Decomposition Transformer with Time-Varying Masked Spatial Attention for Traffic Forecasting\\
}




\author{
\IEEEauthorblockN{Ruiwen Gu}
\IEEEauthorblockA{
\textit{Shenzhen International Graduate School} \\
\textit{Tsinghua University} \\
Shenzhen, China \\
grw23@mails.tsinghua.edu.cn}

\\

\IEEEauthorblockN{Yahao Liu}
\IEEEauthorblockA{
\textit{School of Computer Science and Engineering} \\
\textit{University of Electronic Science and Technology of China} \\
Chengdu, China \\
lyhaolive@gmail.com
}
\and

\IEEEauthorblockN{Qitai Tan}
\IEEEauthorblockA{
\textit{Shenzhen International Graduate School} \\
\textit{Tsinghua University} \\
Shenzhen, China \\
tqt24@mails.tsinghua.edu.cn}

\\

\IEEEauthorblockN{Xiao-Ping Zhang^{*}\thanks{* corresponding author.}}
\IEEEauthorblockA{
\textit{Shenzhen International Graduate School} \\
\textit{Tsinghua University} \\
Shenzhen, China \\
xpzhang@ieee.org}
}

\author{
    \IEEEauthorblockN{%
    Ruiwen Gu\textsuperscript{1}, Qitai Tan\textsuperscript{1}, Yahao Liu\textsuperscript{2}, Xiao-Ping Zhang\textsuperscript{1,*}
}
    \IEEEauthorblockA{\textsuperscript{1} Shenzhen Ubiquitous Data Enabling Key Lab \\
    Shenzhen International Graduate School, Tsinghua University, Shenzhen, China}
    \IEEEauthorblockA{\textsuperscript{2}School of Computer Science and Engineering \\ 
    University of Electronic Science and Technology of China, Chengdu, China}
    \IEEEauthorblockA{\textsuperscript{1} \{grw23, tqt24\}@mails.tsinghua.edu.cn, xpzhang@ieee.org\\ 
    \textsuperscript{2} lyhaolive@gmail.com}
    \thanks{* Corresponding author.}
}



\maketitle

\begin{abstract}
Accurate traffic forecasting is essential for intelligent transportation systems, supporting a wide range of real-world applications. However, it remains challenging due to two key factors:~(1) Traffic series contain heterogeneous temporal patterns, where stable periodic regularities coexist with event-driven fluctuations. Existing methods often treat them within a unified representation, limiting their ability to capture fine-grained temporal dynamics.~(2)Spatial dependencies among nodes are inherently dynamic and sparse, while dense all-pairs attention often introduces redundant interactions and amplifies noise. To address these issues, we propose ADMFormer, an Adaptive-Decomposition Transformer with Time-Varying Masked Spatial Attention. Specifically, ADMFormer first employs a time-node adaptive gating mechanism to decouple traffic signals into dominant regularities and residual fluctuations that vary across time and nodes. A dual-branch temporal module is then designed to separately capture global periodic dependencies and high-frequency irregular variations from these two decomposed components. Furthermore, ADMFormer introduces a time-varying masked spatial attention that sparsifies spatial interactions based on real-time traffic states, thereby effectively preserving dynamic and informative dependencies. Extensive experiments on four real-world datasets demonstrate that ADMFormer achieves state-of-the-art performance.

\end{abstract}


\section{Introduction}
With the rapid development of Intelligent Transportation Systems~(ITS), accurate traffic forecasting has become a fundamental capability, playing a crucial role in congestion control and real-time route planning for urban traffic management. However, traffic flow often exhibits strong non-stationarity and complex spatio-temporal dependencies, making robust forecasting a challenging task.

Recently, deep learning-based methods have been proposed to model the complex dynamics of traffic flow. Most mainstream approaches follow a spatio-temporal learning paradigm, including Spatial-Temporal Graph Neural Networks~(STGNNs)\cite{STGCN,DGCRN,D2STGNN,DMGSTCN,LASTGNN,MTGNN} and Transformer-based architectures\cite{PDFormer,STWave}. Typically, Sequential modules (e.g., RNNs\cite{DCRNN} or temporal convolutions\cite{GWNet}) are employed to capture temporal dynamics, while spatial modules (e.g., diffusion graph convolutions or spatial attention) are adopted to model interactions among traffic nodes.

\begin{figure*}[!t]
    \centering
    \includegraphics[width=\textwidth]{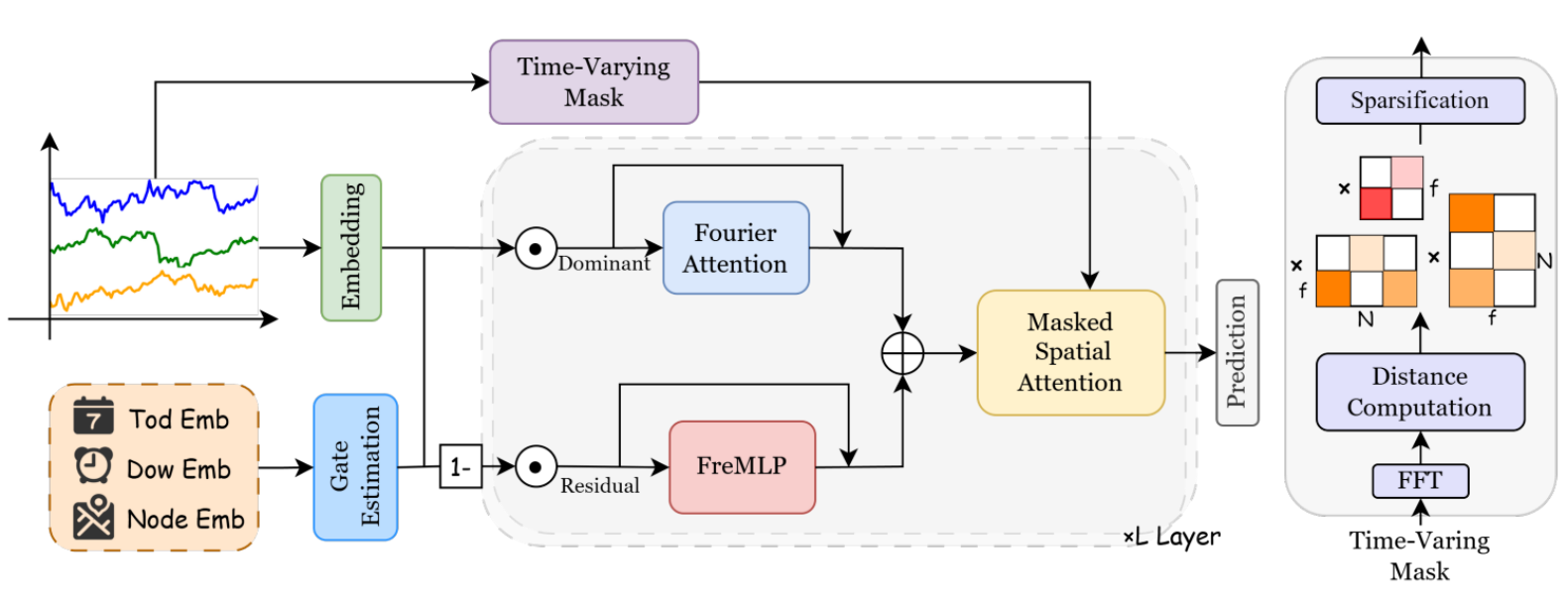}
    \caption{The Framework of ADMFormer and the illustration of Time-varying Mask.}
    \label{fig:admformer}
\end{figure*}

Despite these encouraging results, existing methods still suffer from two major limitations. First, real-world traffic flow exhibits the coexistence of stable periodic patterns~(e.g., daily and weekly human routines) and event-driven irregular fluctuations. These two components arise from different generative mechanisms and exhibit distinct distributions, while their relative importance varies across time slots and nodes. However, many existing methods fail to model such heterogeneous temporal dynamics. They tend to compress diverse temporal patterns into a shared representation space, leading to feature interference. Moreover, traditional decomposition-based methods~(e.g., trend–seasonal decomposition) usually rely on local statistical measures, making it difficult to perform decomposition from a long-term perspective and preventing end-to-end training. Second, spatial dependencies in real road networks are both dynamic and sparse. However, the spatial attention used in prior work typically computes dense all-pairs interactions, introducing irrelevant connections and amplifying noise. In contrast, graph convolution methods often rely on static graph learned from node embeddings, failing to capture evolving traffic interactions. Consequently, the failure to handle both dynamic changes and structural sparsity significantly limits the model's performance on complex road networks.

To address the above limitations, we introduce \textbf{ADMFormer}, an \textbf{A}daptive-\textbf{D}ecomposition Transformer with Time-Varying \textbf{M}asked Spatial Attention. Specifically, ADMFormer consists of two key components.~(1) Adaptive Decomposition for Temporal Dynamics: The model first employs a Time-Node Adaptive Decomposition module that decouples the input traffic signals into two complementary components: time-driven regularities and residual fluctuations. The decomposition is achieved via an adaptive estimation gate conditioned on learnable temporal and node embeddings. Based on the decomposed signals, a Dual-Branch Temporal Learning Module is designed to model each component separately, where Fourier Attention is used to capture global periodic patterns of regularities, and a Complex Fourier Multi-Layer Perceptron~(FreMLP) is adopted to model high-frequency variations in residual fluctuations.~(2) Time-varying Masked Spatial Attention: After fusing temporal representations from both branches, a dynamic spatial mask is generated from real-time inputs and incorporated into spatial attention. The mask is derived using a robust frequency-domain distance metric and sparsified to select informative node interactions in a time-varying manner, thereby enhancing dynamic spatial dependency modeling.

Our main contributions are summarized as follows.

\textbullet \ We propose ADMFormer, an Adaptive-Decomposition Transformer with Time-Varying Masked Spatial Attention for traffic forecasting. which explicitly disentangles mixed temporal patterns and enables time-varying sparse modeling of spatial interactions.

\textbullet \ We design an Adaptive Decomposition module for temporal dynamics, consisting of a Time–Node Adaptive Decomposition module and a Dual-Branch Temporal Learning module. By decoupling traffic signals into regularities and fluctuations and modeling them separately, the proposed design captures fine-grained temporal evolution.

\textbullet \ We introduce a Time-varying Masked Spatial Attention module that generates real-time mask based on the frequency-domain distance metric, which effectively captures dynamic and sparse spatial dependencies.

\textbullet \ We conduct extensive experiments on four real-world datasets. Results demonstrate that our method outperforms state-of-the-art baselines.

\section{Methodology}
The general framework of ADMFormer is illustrated in Fig. \ref{fig:admformer}. Centered on temporal and spatial modeling, the model consists of four modules. We present the technical details of these modules in this section.

\subsection{Data Embedding Module}
The Data Embedding Module enhances the model's representation by integrating three types of information: raw observations, temporal priors and spatial identities. Given the historical traffic flow $X \in \mathbb{R}^{T \times N \times C}$, where $T$, $N$, and $C$ denote the sequence length, the number of traffic nodes, and the feature dimension, respectively. The raw observations are first projected into the embedding space through a multilayer perceptron: $X_{emb} = \mathrm{MLP}(X)\in \mathbb{R}^{T \times N \times D}$, where $D$ denotes the embedding dimension.

Traffic data usually exhibits strong daily and weekly periodicity. To encode such temporal priors, we introduce two learnable time-slot embeddings $T^{d}\in\mathbb{R}^{288\times D}$ and $T^{w}\in\mathbb{R}^{7\times D}$ (for 5-minute data, one day has 288 slots and one week has 7 days). For each timestep $t$, the corresponding time embeddings are retrieved according to its Time-of-Day index $i \in \{1, \dots, 288\}$ and Day-of-Week index $j \in \{1, \dots, 7\}$, yielding $T^d[i]$ and $T^w[j]$. The time embeddings over the past $T$ steps are then stacked to form $E^d$ and $E^w \in \mathbb{R}^{T \times D}$

In addition, different nodes have unique spatial properties. Thus, we assign each node a learnable node embedding $E^N \in \mathbb{R}^{N \times D}$ to capture such static spatial information. The above three embeddings will be ultilized in the subsequent modules.

\subsection{Time-Node Adaptive Decomposition Module}
The Time-Node Adaptive Decomposition Module is designed to disentangle the embedded traffic  representations into two complementary components: a dominant component that captures node-specific periodic patterns and a residual component that models irregular fluctuations.

To adaptively estimate the proportion of the dominant component, the module leverages prior learnable temporal and spatial embeddings. Specifically, the Tod embedding $E^{d}$, DoW embedding $E^{w}$, and node embedding $E_N$ are first broadcast to $\mathbb{R}^{T\times N \times D}$ and concatenated along the feature dimension. The concatenated representation is then fed into a lightweight MLP parameterized by $W_g \in \mathbb{R}^{3D \times D}$ and $b_g \in \mathbb{R}^{D}$, followed by a sigmoid activation to generate gating values $\Lambda \in \mathbb{R}^{T\times N \times D}$ in $[0,1]$:
\begin{equation}
\Lambda=\sigma(W_g(E^d \Vert E^w \Vert E^N)+b_g)\label{eq}
\end{equation}
Where $\Vert$ denotes concatenation and $\sigma(\cdot)$ is the sigmoid function. The dominant component $X_{\mathrm{main}}$ is then derived through element-wise product between the gating values $\Lambda$ and embedded traffic features $X_{emb}$, while the residual component $X_{\mathrm{res}}$ is defined as the remainder:
\begin{equation}
X_{\mathrm{main}} = X_{emb} \odot \Lambda,\qquad
X_{\mathrm{res}} = X_{emb} \odot (1-\Lambda)
\end{equation}

By generating time–node conditioned gating values, the module adaptively decomposes traffic representations for each node and timestep, supporting subsequent dual-branch temporal learning on the two disentangled components.

\subsection{Dual-Branch Temporal Learning Module}
To model the distinct temporal patterns of the decomposed components, we design a Dual-Branch Temporal Learning Module.

For the dominant part $X_{\mathrm{main}}$, we employ Fourier Attention to capture global periodic correlations. By operating in the frequency domain, the model can directly capture periodic dependencies through dominant spectral components while remaining robust to local spikes. In detail, the query, key, and value are first generated from $X_{\mathrm{main}}$, followed by the Fast Fourier Transform (FFT) to obtain their spectral representations. After computing the attention scores, the output will be transformed back via the inverse Fast Fourier Transform~(IFFT).
\begin{equation}
\left\{
\begin{aligned}
Q_F &= \mathrm{FFT}\!\left(W_q X_{main}\right),\\
K_F &= \mathrm{FFT}\!\left(W_k X_{main}\right),\\
V_F &= \mathrm{FFT}\!\left(W_v X_{main}\right).
\end{aligned}
\right.
\end{equation}

\begin{equation}
\hat{X}_{\mathrm{main}}
= \mathrm{iFFT}\!\Big(\mathrm{softmax}\!\left(\frac{Q_F K_F^{\mathrm{H}}}{\sqrt{d}}\right)V_F\Big),
\end{equation}
Where $d$ denotes the dimensionality of each attention head. The residual component $X_{\mathrm{res}}$ consists of irregular, non-periodic fluctuations characterized by high-frequency variations. To address this, we employ a Complex Fourier Multi-Layer Perceptron~(FreMLP). Specifically, $X_{\mathrm{res}}$ is first mapped to the frequency domain via FFT, where a complex-valued MLP is applied to learn the underlying dynamics from the spectral representations.
\begin{equation}
\hat{X}_{\mathrm{res}}
= \mathrm{iFFT}\!\Big( \mathrm{FreMLP}\big( \mathrm{FFT}(X_{\mathrm{res}}) \big) \Big)
\end{equation}

In detail, the Complex MLP processes the complex-valued spectral coefficients by separately updating their real and imaginary parts. This allows the model to simultaneously capture amplitude and phase variations, formulated as follows:
\begin{equation}\label{eq:Fremlp}
\begin{aligned}
\Re(\hat{X}_\mathrm{res}^F) &= ReLU\!\left(\Re(X_\mathrm{res}^F)W_{r}-\Im(X_\mathrm{res}^F)W_{i}+b_{r}\right),\\
\Im(\hat{X}_{res}^F) &= ReLU\!\left(\Im(X_\mathrm{res}^F)W_{r}+\Re(X_\mathrm{res}^F)W_{i}+b_{i}\right).
\end{aligned}
\end{equation}
Where $W = W_{r}+jW_{i}$, $b=b_{r}+jb_{i}$ are complex learnable parameters. Finally, the outputs of the two temporal branches are integrated to form a comprehensive representation of the traffic flow:  $\hat{X}=\hat{X}_\mathrm{main}+\hat{X}_\mathrm{res}$, and then fed into the subsequent spatial learning module.

\subsection{Time-Varying Masked Spatial Attention Module}
The Time-Varying Masked Spatial Attention is designed to model dynamic dependencies while maintaining structural sparsity. Instead of using predefined distances or costly DTW preprocessing, our approach dynamically generates input-adaptive masks in the frequency domain. These masks act as filters to refine spatial interactions and reduce noise.

In the mask computation process, we employ a learnable quadratic distance in the frequency domain as a robust similarity metric that is less sensitive to noise. Specifically, we first transform the input $X$ into $X^F$ via FFT and compute the pairwise differences between nodes. A symmetric learnable matrix $W_d$ is then introduced to define a quadratic form over these differences, thereby enhancing the model’s expressiveness. Since larger distances indicate weaker functional relationships between nodes, the continuous connectivity score $S_{ij}$ is defined using the inverse of the modulated distance, as follows:
\begin{equation}\label{eq:Dij_inv}
S_{ij}=\mathrm{}\frac{1}{\left(X^{F}_{i,:}-X^{F}_{j,:}\right)W_d \left(X^{F}_{i,:}-X^{F}_{j,:}\right)^{T} +\epsilon}.
\end{equation}
where $\epsilon$ is a small constant introduced for numerical stability. The connectivity scores are then max-normalized into a probability matrix and discretized into a sparse binary mask $M_{ij} \sim \mathrm{Bernoulli}(S_{ij}) \in \{0,1\}^{N \times N}$ via Bernoulli resampling. To make this sampling trainable, we employ the Gumbel-Sigmoid\cite{duet} reparameterization, which makes the discrete sampling process differentiable.

Finally, the learnable mask is incorporated into the spatial attention computation via element-wise operation. This guides the model to focus on the most informative spatial interactions, following as follows:
\begin{equation}\label{eq:spa_attn}
\begin{gathered}
A_s=\mathrm{Softmax}\!\left(
\frac{Q_s K_s^{\top}}{\sqrt{d_m}} \odot \mathcal{M}
+\left(1-\mathcal{M}\right)\odot(-\infty)
\right),\\
O_s=A_sV_s .
\end{gathered}
\end{equation}
where $Q_s$, $K_s$, and $V_s$ are the query, key, and value computed from temporal representation $\hat{X}$. This module enables the model to adaptively adjust spatial connectivity, balancing dynamic interactions and structural sparsity.

\subsection{Output Layer}
The final representations $O_s$ are mapped to the future $H$-step predictions $\hat{Y} \in \mathbb{R}^{H \times N}$ via a linear projection layer.

To optimize the model, we adopt the Huber Loss as the training objective, defined as follows:
\begin{equation}\label{eq:huber_loss}
\mathcal{L}=\frac{1}{H \cdot N}\sum_{i=1}^{H}\sum_{j=1}^{N}
\ell_{\delta}\!\left(Y^{gt}_{i,j}-\hat{Y}_{i,j}\right),
\end{equation}
where $Y^{gt}$ is the ground truth, $\ell_{\delta}$ is defined as:
\begin{equation}
\ell_{\delta}(x)=
\begin{cases}
\frac{1}{2}x^2, & |x|\le \delta,\\
\delta\left(|x|-\frac{1}{2}\delta\right), & |x|>\delta,
\end{cases}
\end{equation}
where $\delta$ is a preset threshold parameter.

\section{Experiments}

In this section, we evaluate ADMFormer on real-world traffic datasets, including experimental settings, comparisons with SOTA methods, ablation and sensitivity studies, and visualization results.

\begin{table}[!b]
\caption{Datasets Description.}
\begin{center}
\begin{tabular}{|c|c|c|c|}
\hline
\textbf{Datasets} & \textbf{Node} & \textbf{Time step} & \textbf{Time Range} \\
\hline
PeMS03 & 358 & 26202 & 09/01/2018--11/30/2018 \\
\hline
PeMS04 & 307 & 16992 & 01/01/2018--02/28/2018 \\
\hline
PeMS07 & 883 & 28224 & 05/01/2017--08/31/2017 \\
\hline
PeMS08 & 170 & 17856 & 07/01/2016--08/31/2016 \\
\hline
\end{tabular}
\label{tab:datasets}
\end{center}
\end{table}

\subsection{Experiment Settings}\label{AA}
\textbullet \ \noindent\textbf{\textit{Datasets:}} We evaluate the performance of ADMFormer on four real-world traffic datasets: PeMS03, PeMS04, PeMS07, and PeMS08. These datasets were collected from the California Performance Measurement System (PeMS), with detailed statistics provided in Table \ref{tab:datasets}. All datasets are sampled at 5-minute intervals, resulting in 12 time steps per hour.

\textbullet \ \noindent\textbf{\textit{Baselines:}} We compare ADMFormer with  the following 10 baselines, including traditional statistical methods, classic deep learning  methods and the recent SOTA models, as listed below: ARIMA\cite{ARIMA}, STGCN\cite{STGCN}, DCRNN\cite{DCRNN}, GWNet\cite{GWNet}, DGCRN\cite{DGCRN}, GMAN\cite{GMAN}, PDFormer\cite{PDFormer}, D$^2$STGNN\cite{D2STGNN}, STWave\cite{STWave} and STDN\cite{STDN}.

\textbullet \ \noindent\textbf{\textit{Evaluation Metrics:}} We employ three standard metrics to evaluate the performance of different methods: Mean Absolute Error (MAE), Mean Absolute Percentage Error (MAPE), and Root Mean Squared Error (RMSE).

\textbullet \ \noindent\textbf{\textit{Parameter Settings:}} In line with prior research, all four datasets are split into training, validation, and test sets with a ratio of 6:2:2. Normalization is used to stabilize the input data. For short-term forecasting, both the input length $T$ and the prediction horizon $H$ are set to 12 time steps, with the input dimension fixed at 3.

All experiments are conducted on a server with 2 NVIDIA GeForce RTX 4090 GPUs. The model is trained using the Adam optimizer with an initial learning rate of 0.001. And it is trained for up to 100 epochs, using multi-step learning rate decay and an early stopping strategy with a patience of 20 to prevent overfitting. The batch size is fixed at 16. The number of layers $L$ and embedding dimension $D$ are tuned within $\{1,2,3,4\}$ and $\{32,64,96,128\}$ respectively. The number of attention heads is fixed at 4.


\begin{table*}[t]
\caption{Performance Comparison on PEMS datasets.}
\begin{center}
\scriptsize
\renewcommand{\arraystretch}{1.15}
\setlength{\tabcolsep}{3.5pt}

\resizebox{\textwidth}{!}{%
\begin{tabular}{|c|ccc|ccc|ccc|ccc|}
\hline
\textbf{Method} &
\multicolumn{3}{c|}{\textbf{PEMS03}} &
\multicolumn{3}{c|}{\textbf{PEMS04}} &
\multicolumn{3}{c|}{\textbf{PEMS07}} &
\multicolumn{3}{c|}{\textbf{PEMS08}} \\
\cline{2-13}
 & \textbf{MAE} & \textbf{RMSE} & \textbf{MAPE} &
   \textbf{MAE} & \textbf{RMSE} & \textbf{MAPE} &
   \textbf{MAE} & \textbf{RMSE} & \textbf{MAPE} &
   \textbf{MAE} & \textbf{RMSE} & \textbf{MAPE} \\
\hline

\hline
ARIMA\cite{ARIMA} & 35.41 & 47.59 & 33.78\% & 33.73  & 48.80 & 24.18\% & 38.17 & 59.27 & 19.46\% & 31.09 & 44.32 & 22.73\% \\
STGCN\cite{STGCN}  & 17.49 & 30.12 & 17.15\% & 22.70 & 35.55 & 14.59\% & 25.38 & 38.78 & 11.08\% & 18.02 & 27.83 & 11.40\% \\
DCRNN\cite{DCRNN}  & 18.18 & 30.31 & 18.91\% & 24.70 & 38.12 & 17.12\% & 25.30 & 38.58 & 11.66\% & 17.86 & 27.83 & 11.45\% \\
GWNet\cite{GWNet}  & 19.85 & 32.94 & 19.31\% & 25.45 & 39.70 & 17.29\% & 26.85 & 42.78 & 12.12\% & 19.13 & 31.05 & 12.68\% \\
GMAN\cite{GMAN} & 16.87 & 27.92 & 18.23\% & 19.14 & 31.60 & 13.19\% & 20.97 & 34.10 & 9.05\% & 15.31 & 24.92 & 10.13\% \\
DGCRN\cite{DGCRN}  & \underline{14.74} & 25.97 & 15.42\% & 18.73 & 30.65 & 12.82\% & 20.34 & 33.22 &  8.84\% & 14.30 & 23.50 &  9.33\% \\
PDFormer\cite{PDFormer} & 14.94 & \textbf{25.39} & 15.82\% & \underline{18.32} & \underline{29.96} & \underline{12.10}\% & 19.83 & \underline{32.87} & 8.53\% & 13.58 & 23.51 & 9.04\% \\
D$^2$STGNN\cite{D2STGNN} & 15.10 & 26.57 & 15.23\% & 18.42 & 29.97 & 12.81\% & \underline{19.68} & 33.24 & 8.43\% & 14.35 & 24.18 & 9.33\% \\
STWave\cite{STWave} & 14.93 & 26.50 & \underline{15.05}\% & 18.50 & 30.39 & 12.43\% & 19.94 & 33.88 & \underline{8.38}\% & \underline{13.44} & \underline{23.36} & \underline{8.94}\% \\
STDN\cite{STDN}  & 16.05 & 27.51 & 17.71\% & 18.67 & 30.92 & 13.16\% & 22.94 & 36.06 & 10.32\% & 14.79 & 24.60 & 10.26\% \\
\hline

\textbf{ADMFormer} &
\textbf{14.44} & \underline{25.94} & \textbf{14.58}\% &
\textbf{18.28} & \textbf{29.90} & \textbf{11.98}\% &
\textbf{18.90} & \textbf{32.33} & \textbf{7.87}\% &
\textbf{13.41} & \textbf{22.85} & \textbf{8.83}\% \\
\hline
\end{tabular}%
}
\label{tab:pems_main_noLeftCol}
\end{center}
\end{table*}

\subsection{Performance Comparison}
The performance comparison on traffic flow datasets is presented in Table \ref{tab:pems_main_noLeftCol}, where the best results are highlighted in bold and the second best are underlined. The results lead to the following observations:

1) Overall, ADMFormer achieves the best performance. Notably, our model achieves significant improvements of 3.96\%, 2.18\%, and 6.09\% in MAE, RMSE and MAPE, respectively.
2) Classical statistical methods generally exhibit inferior performance, as they fail to model nonlinear temporal dynamics and neglect spatial correlations.
3) Attention-based models, such as GMAN and PDFormer, exhibit superior performance by capturing long-range spatial correlations. However, their spatial attention densely computed and lacks explicit time-varying constraints, which may introduce noisy or irrelevant dependencies.
4) Decoupling-based methods, such as STWave and D$^2$STGNN, achieve better results through decomposition. However, their decomposition are typically static and insufficient to fully disentangle complex temporal dependencies. 
Compared with them, our ADMFormer improves performance through adaptive decomposition guided by spatio-temporal priors, dual-branch temporal modeling, and time-varying spatial constraints within spatial attention, enabling more effective spatio-temporal dependency modeling.

\subsection{Ablation Study}

\begin{figure}[htbp]
    \centering
    \includegraphics[width=\linewidth]{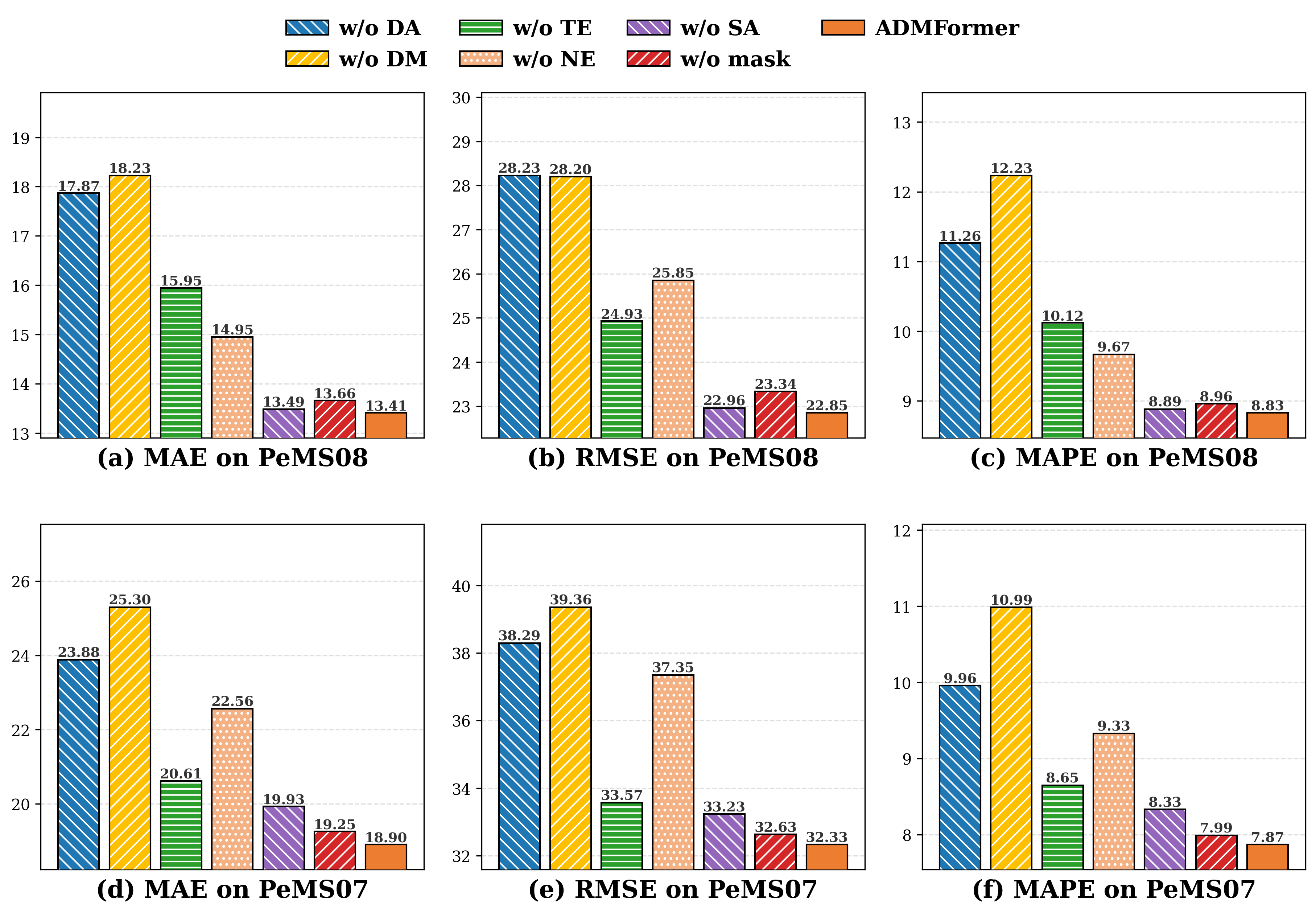}
    \caption{Ablation study on PEMS07 and PeMS08.}
    \label{fig:ablation}
\end{figure}

To further investigate the contribution of each component in our model, we conduct an ablation study on the following variants of ADMFormer:
1) w/o DA: it removes the adaptive decomposition module and uses Fourier Attention for temporal modeling.
2) w/o DM: it removes the adaptive decomposition module and uses FreMLP for temporal modeling.
3) w/o NE: it removes node embedding in gating generation. 
4) w/o TE: it removes time embeddings in gating generation. 
5) w/o SA: it removes the spatial attention module 
6) w/o mask: it removes the time-varying mask.

Figure \ref{fig:ablation} shows the ablation results of ADMFormer on the PEMS07 and PeMS08 datasets. From these results, we can draw the following conclusions: 

1) w/o DA and w/o DM exhibit the two poorest performances, highlighting the necessity of the Adaptive Decomposition Module. Without it, the model fails to effectively disentangle regular patterns and fluctuations, and also loses critical guidance from temporal and spatial priors.
2) The performance drops of w/o NE and w/o TE show the importance of both spatial and temporal priors in gate generation. These priors enable the decomposition to adapt to node and time-specific features, effectively capturing data heterogeneity.
3) w/o SA causes a clear performance degradation on PeMS07, indicating that spatial dependencies are crucial for this dataset. But interestingly, it outperforms the full spatial attention variants on PeMS08. This suggests that standard full attention may introduce noisy correlations and hurt prediction. However, our proposed masked attention suppresses such noisy connections, thereby enhancing predictive accuracy. 
4) The performance of w/o mask is worse than the full model, which demonstrates that our learned mask is effective.

\subsection{Parameter Sensitivity Analysis}

\begin{figure}[htbp]
    \centering
    \includegraphics[width=\linewidth]{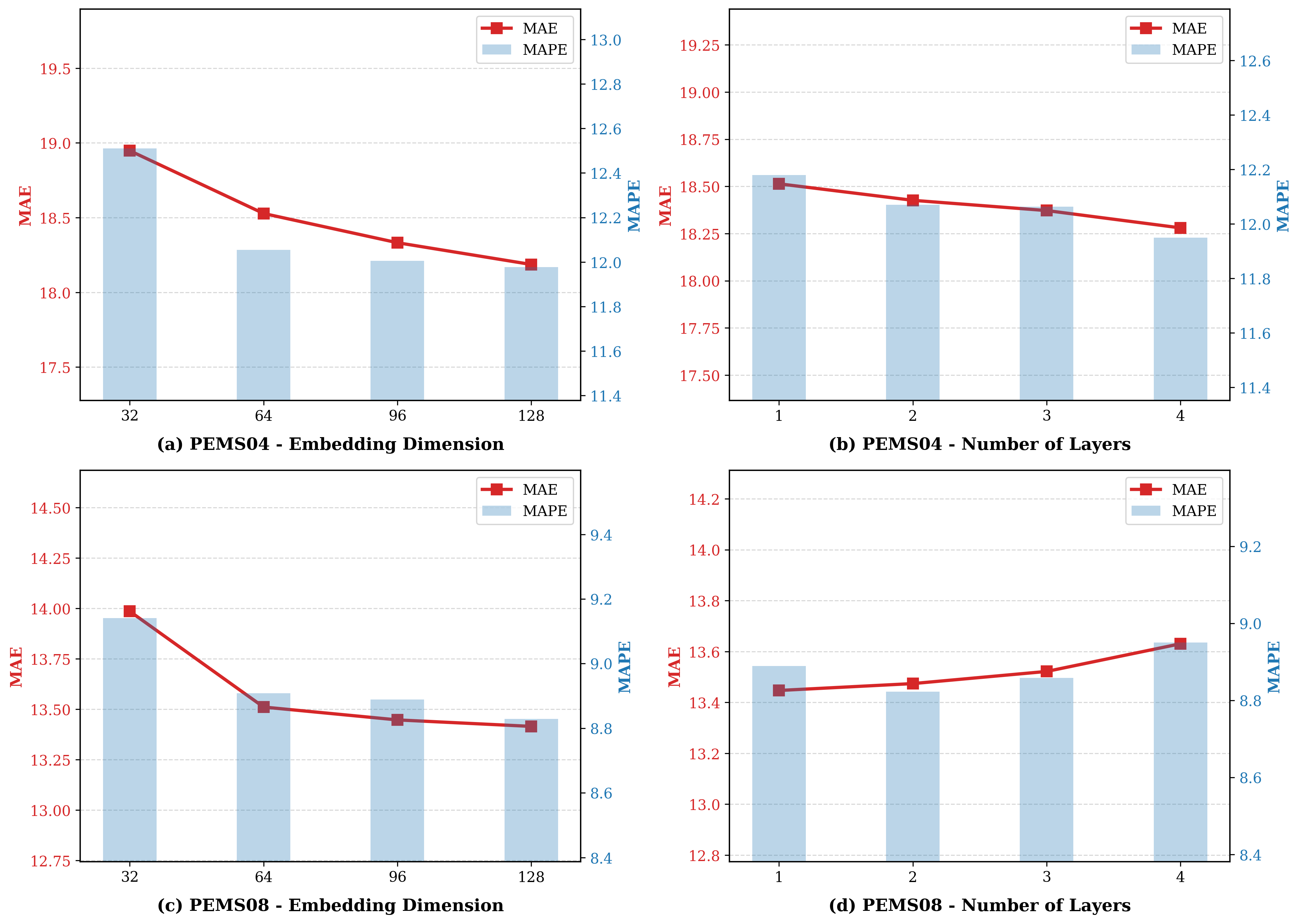}
    \caption{Parameter sensitivity analysis on PEMS04 and PeMS08.}
    \label{fig:sensitivity}
\end{figure}

Fig.~\ref{fig:sensitivity} shows the parameter sensitivity analysis of ADMFormer on the PeMS04 and PeMS08 datasets. We evaluate two key hyperparameters: the number of model layers $L$ and the embedding dimension $D$, with $L$ explored in $\{1, 2, 3, 4\}$ and $D$ explored in $\{32, 64, 96, 128\}$. The results lead to the following two observations.

1) The embedding dimension $D$ significantly influences the representation capacity of the model. For both PeMS04 and PeMS08, increasing $D$ consistently improves performance, with the best results achieved at $D=128$ within the explored range. This indicates that richer embeddings help the model capture complex spatio-temporal patterns.
2) The number of layers $L$ is sensitive to dataset complexity. On PeMS04, increasing $L$ leads to steadily improved performance, indicating that additional layers help capture more complex dependencies. In contrast, on PeMS08, performance degrades with increasing $L$, implying a higher risk of overfitting on the smaller dataset.

\begin{figure*}[htbp]
    \centering
    \includegraphics[width=\textwidth]{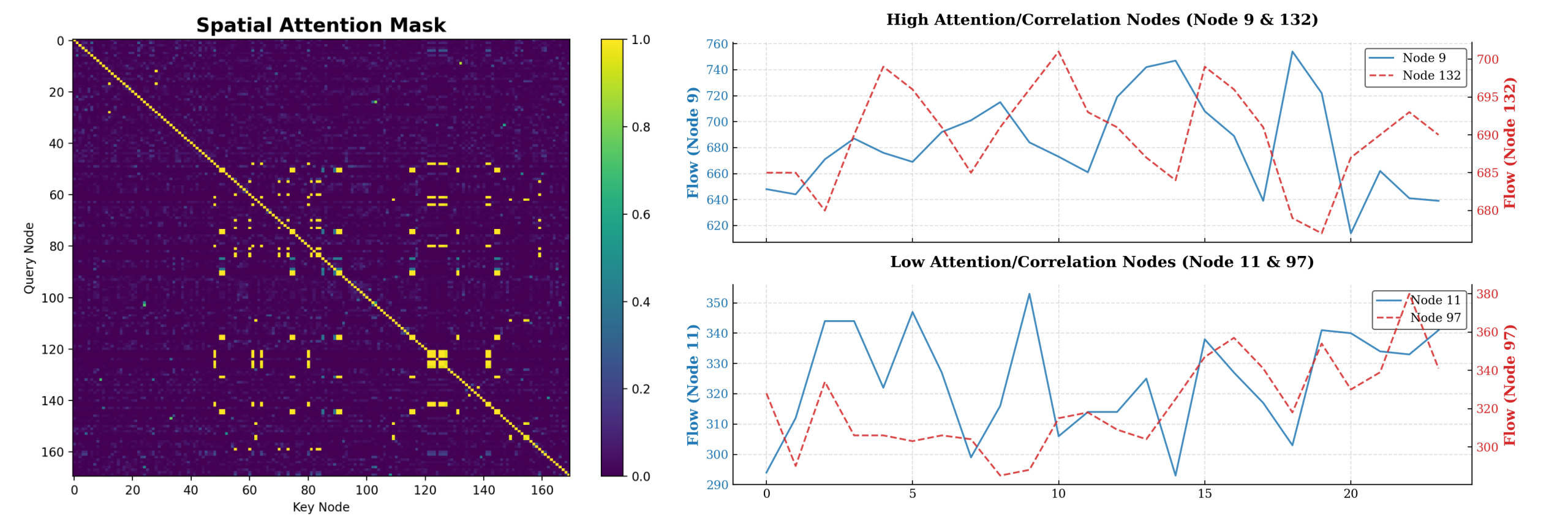}
    \caption{Analysis of the Frequency-aware Mask: Heatmap of the learned mask (left) and traffic flow visualization between node pairs (right).}
    \label{fig:mask_analysis}
\end{figure*}

\subsection{Visualization}
To evaluate the effectiveness of the time-varying masking mechanism, we visualize the learned mask and compare the traffic pattern of node pairs with different mask values. 

As shown in Fig.~\ref{fig:mask_analysis} (left), the learned mask exhibits a clear sparse structure, where most node pairs are assigned low values (adarker regions), while only a small subset of node pairs shows strong connectivity (bright regions). This demonstrates that the computed mask effectively filters redundant interactions and focuses on the most informative spatial dependencies. 

To further analyze the interpretability of the learned mask, we select two pairs of nodes with different mask values and plot their traffic flow. As shown in Fig.~\ref{fig:mask_analysis} (right), node pairs with high mask values (e.g., nodes 9 and 132) exhibit highly correlated temporal patterns, indicating a strong functional correlation. In contrast, nodes with low mask values (e.g., nodes 11 and 97) show weak and distinct patterns. This suggests that our proposed time-varying masking mechanism effectively captures intrinsic spatial dependencies in the real-world road network.

\section{Conclusion}
In this paper, we propose ADMFormer, an Adaptive-Decomposition Transformer with Time-Varying Masked Spatial Attention for traffic forecasting. To address two core challenges in real-world road networks: entangled temporal dynamics and the lack of time-varying sparsity in spatial modeling. The framework follows a decoupling-based learning paradigm. Specifically, a Time-Node Adaptive Decomposition module is first employed to decouple raw signals into dominant regularities and residual fluctuations. A Dual-Branch Temporal Learning module is then developed to model the distinct temporal dependencies of the two decomposed components. For spatial modeling, we introduce a Time-Varying Masked Spatial Attention mechanism that generates dynamic sparse masks from real-time inputs based on frequency-domain distances, effectively filtering redundant interactions. Extensive experiments on four real-world traffic datasets show that ADMFormer significantly improves prediction accuracy, offering a new perspective for the future design of spatio-temporal Transformers.




\section*{Acknowledgment}
This work is supported by Shenzhen Ubiquitous Data Enabling Key Lab under grant ZDSYS20220527171406015. and by Tsinghua Shenzhen International Graduate School-Shenzhen Pengrui Endowed Professorship Scheme of Shenzhen Pengrui Foundation.

\bibliographystyle{IEEEtran} 
\bibliography{references}    

\end{document}